\documentclass[conference]{IEEEtran}
\IEEEoverridecommandlockouts
\usepackage{cite}
\usepackage{amsmath,amssymb,amsfonts}
\usepackage{algorithmic}
\usepackage{graphicx}
\usepackage{textcomp}
\usepackage{xcolor,soul,framed}
\usepackage{epstopdf}
\usepackage{float}
\usepackage{subcaption}
\usepackage{balance}
\newdimen\figrasterwd
\figrasterwd\textwidth

\def\BibTeX{{\rm B\kern-.05em{\sc i\kern-.025em b}\kern-.08em
    T\kern-.1667em\lower.7ex\hbox{E}\kern-.125emX}}
\begin{document}

\title{Histogram Layers for Synthetic Aperture Sonar Imagery\\
\thanks{This material is based upon work supported by the National Science Foundation Graduate Research Fellowship under Grant No. DGE-1842473 and by the Office of Naval Research Grant N00014-16-1-2323. The views and opinions of authors expressed herein do not necessarily state or reflect those of the United States Government or any agency thereof. Distribution Statement A: Approved for public release: distribution is unlimited.}
}

 \author{\IEEEauthorblockN{Joshua Peeples$^1$, Alina Zare$^2$, Jeffrey Dale$^{3,4}$, James Keller$^3$}
 \IEEEauthorblockA{$^1$Department of Electrical and Computer Engineering, Texas A\&M University, College Station, TX, USA\\
	\IEEEauthorblockA{$^2$Department of Electrical and Computer Engineering, University of Florida, Gainesville, FL, USA\\
		\IEEEauthorblockA{$^3$Department of Computer Science and Electrical Engineering, University of Missouri, Columbia, MO, USA\\
    	\IEEEauthorblockA{$^4$Naval Surface Warfare Center Panama City Division, 110 Vernon Ave, Panama City, FL 32407, USA\\
		jpeeples@tamu.edu, azare@eng.ufl.edu, jjdale@mail.missouri.edu, KellerJ@missouri.edu}}}}}
		

\maketitle

\begin{abstract}
Synthetic aperture sonar (SAS) imagery is crucial for several applications, including target recognition and environmental segmentation. Deep learning models have led to much success in SAS analysis; however, the features extracted by these approaches may not be suitable for capturing certain textural information. To address this problem, we present a novel application of histogram layers on SAS imagery. The addition of histogram layer(s) within the deep learning models improved performance by incorporating statistical texture information on both synthetic and real-world datasets.
\end{abstract}


\section{Introduction}
    Synthetic aperture sonar (SAS) produces high resolution images of the seafloor \cite{hayes2009synthetic}. The imagery generated through SAS can be used to perform important tasks such as automatic target recognition \cite{richard2021deep}, environmental segmentation \cite{sun2022iterative}, and automated scene understanding \cite{stewart2021image}. Machine learning methods have been applied to SAS imagery to improve the accuracy and efficiency for these different applications. A critical aspect of the machine learning methods is the representation (\textit{i.e.}, features) of the SAS imagery. As a result, several works investigate feature extraction approaches using a) handcrafted \cite{williams2009unsupervised} and/or b) deep learning \cite{kohntopp2017seafloor} features. Deep learning models, particularly convolutional neural networks (CNNs), have led to great strides in research for processing SAS imagery \cite{richard2021deep,stewart2021image,sun2022iterative,kohntopp2017seafloor}. A main reason for this success is the ability of deep learning frameworks to automate the feature extraction process and perform follow-on tasks (\textit{e.g.}, classification, segmentation). Literature has suggested that CNNs are biased towards texture features rather than shape \cite{geirhos2018imagenet,hermann2020origins}. For SAS-related efforts, texture information is crucial to represent the data \cite{williams2009unsupervised}. CNNs are biased towards a particular type of texture (\textit{i.e.}, \textit{structural} texture) \cite{peeples2021histogram,zhu2021learning}. To improve the information extracted within these deep learning models, other types of texture features can be incorporated into the network.
    
\begin{figure}[t]
	\centering
	\includegraphics[width=.80\linewidth]{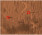}
	\caption{SAS image example for understanding the environmental context that can be used in different tasks (\textit{e.g.}, object recognition). The seafloor changes between different textures types and texture features can help capture this information. The objects of interest in the image are marked by red rectangles.}
	\label{fig:sonar_ex}
	\end{figure}

	Texture information is a common type of image feature that has led to much success \cite{liu2019bow}. For example, texture can be used to provide environmental context  in sonar imagery \cite{peeples2019comparison,zare2017possibilistic,williams2015fast} that can be used in various tasks such as identifying potential hazards. In Figure \ref{fig:sonar_ex}, there are several difficult aspects of the environment that texture features can capture. For example, the seafloor types blend (\textit{e.g.}, sand ripple patterns change, overlap between different seafloor types) and texture approaches can model these gradients and regions of transition that occur within the SAS imagery.
	Texture is always present within images \cite{haralick1973textural}; therefore, the representation of texture content within an image is an important component of texture analysis. Additionally, the need for informative, domain-specific (\textit{e.g.}, SAS) texture information has led to several methods for texture feature extraction, particularly histogram-based approaches \cite{zhu2014model,peeples2019comparison,cobb2010parametric}.
	
\begin{figure}[t]
	\centering 
	\hspace{-14mm}
	\includegraphics[width=1\linewidth]{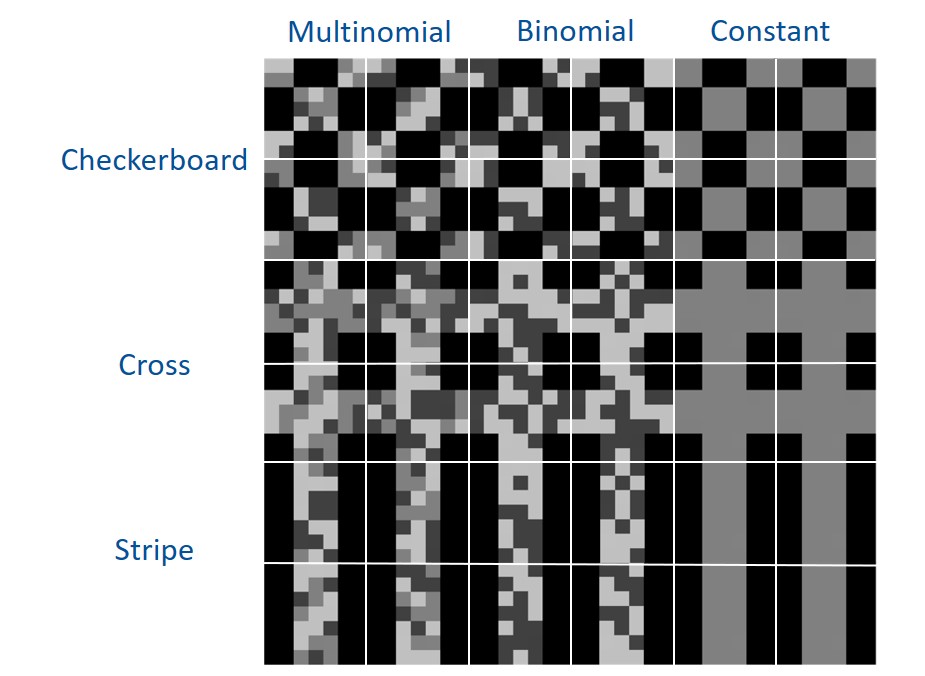}
	\caption{Examples of different types of structural and statistical textures. The structural and statistical textures vary along the vertical and horizontal axis respectively.}
	\label{fig:textures}
	\end{figure}

	As a result of the success of histogram-based, handcrafted features, histogram layer(s) have been proposed \cite{peeples2021histogram,sedighi2017histogram,wang2016learnable,yusuf2020differentiable,sadeghi2022histnet}. These methods bridge the gap between handcrafted and deep learning approaches to improve the \textit{statistical} texture context within the model. Structural textures consist of defining a set of texture examples and an order of spatial positions for each exemplar \cite{materka1998texture}. On the other hand, statistical textures represent the data through parameters that characterize the distributions and correlation between the intensity and/or feature values in an image, as opposed to understanding the structure of each texture \cite{humeau2019texture}. Methods that focused on extracting statistical texture, particularly those computing measures between pairs of pixels (\textit{i.e.}, second order), have outperformed the other texture analysis approaches \cite{weszka1976comparative,haralick1979statistical,ojala1996comparative}. To illustrate the difference between structural and statistical textures, we show an example in Figure \ref{fig:textures}. The structural textures are a checkboard, cross, and stripe displayed along the vertical axis of the figure. The statistical textures are shown along the horizontal axis where the foreground pixels values were sampled from a multinomial, binomial, and constant distributions. CNNs would have difficulty learning weights that could account for both the spatial ordering of the pixels (\textit{i.e.}, structural changes) and individual pixel changes (\textit{i.e.}, statistical changes). However, histogram layer(s) could be used to learn the statistical distributions to further enhance the features learned by the model to correctly classify the textures \cite{peeples2021histogram}.
	
	Inspired by the utility of statistical texture features, histogram layer(s) can be used to provide a powerful representation of the data more suitable for SAS imagery. We hypothesize that incorporating histogram layer(s) within deep learning models will improve performance on SAS-related tasks. The contributions of this work are the following:
	\begin{itemize}
	    \item First application of histogram layer(s) for SAS imagery
	    \item Investigation into model configuration for histogram layer(s)
	    \item Thorough analysis of statistical and structural textures for SAS imagery
	\end{itemize}

\section{Method}
\subsection{Histogram Layer Review}
We briefly summarize the local, radial basis function (RBF) histogram layer that was previously introduced \cite{peeples2021histogram}. 
The histogram layer can be applied to several data modalities with spatial information (\textit{e.g.}, grayscale, RGB, and convolutional feature maps). Histograms generally require three hyperparameters: bin centers ($\gamma_{bd}$), bin widths ($\mu_{bd}$), and the number of bins ($B$). With the histogram layer, the bin centers and widths are learned by the network instead of setting these values \textit{a priori}. Given input data, $\mathbf{X} \in \mathbb{R}^{M \times N \times D}$, where $M$ and $N$ are the spatial dimensions while $D$ is the feature dimensionality, the output tensor of the local histogram layer, $\mathbf{Y} \in \mathbb{R}^{R \times C \times B \times D}$ with spatial dimensions $R$ and $C$ after applying a histogram layer with kernel size $S \times T$ is shown in Equation \ref{eqn:binning}:
\begin{equation}
    Y_{rcbd} =  
	\cfrac{1}{ST}\sum_{s=1}^{S}\sum_{t=1}^{T}e^{-\gamma_{bd}^2\left(x_{r+s,c+t,d}-\mu_{bd}\right)^2}.
	\label{eqn:binning}
\end{equation}
Each input feature dimension is treated independently; therefore, $BD$ histogram features maps are extracted (\textit{e.g.}, three input channels and three bins would result in nine histogram feature maps). The histogram layer is constructed using pre-existing layers that add flexibility and ease of implementation.

\subsection{Network Architectures}
\begin{figure*}[htb]
	\centering
	\parbox{\figrasterwd}{
		\parbox{.545\figrasterwd}{%
			\subcaptionbox{Resnet18 Backbone}{\includegraphics[width=\hsize]{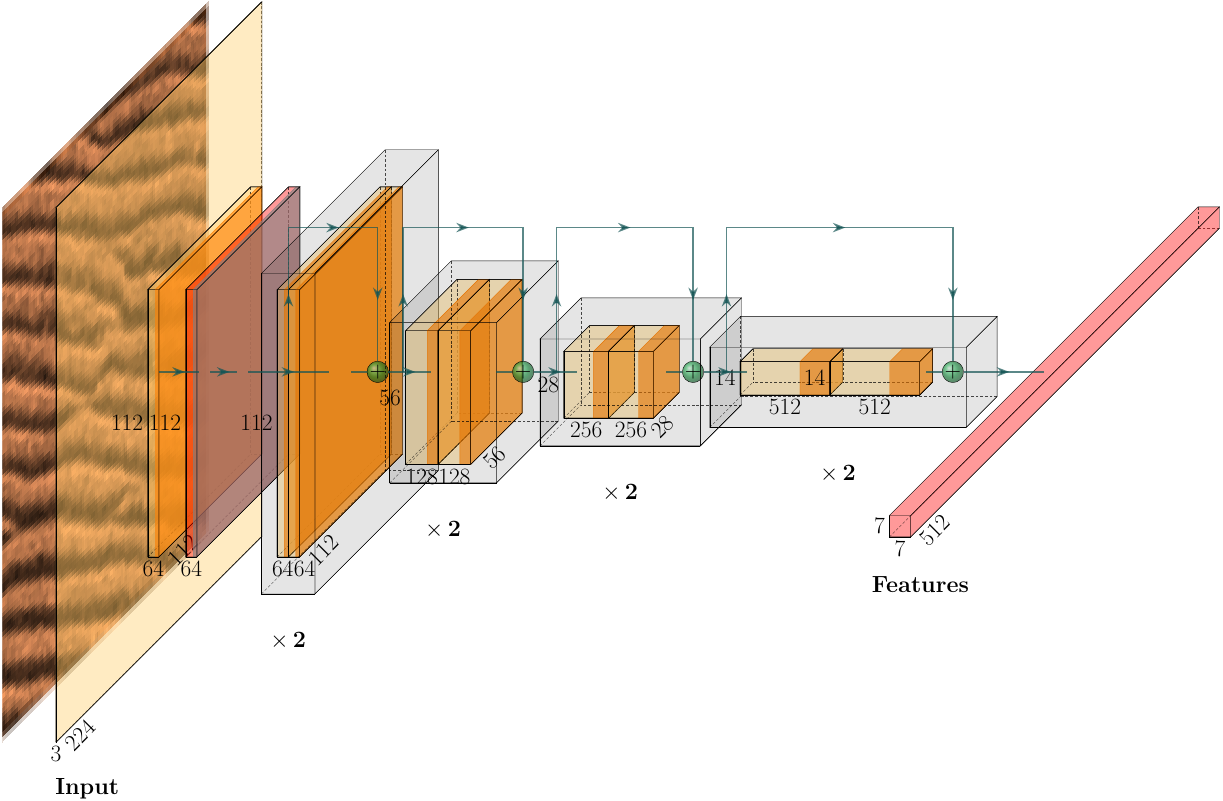}\label{fig:backbone}}
		}
	\centering
		\parbox{.44\figrasterwd}{%
			\subcaptionbox{Parallel}{\includegraphics[width=\hsize]{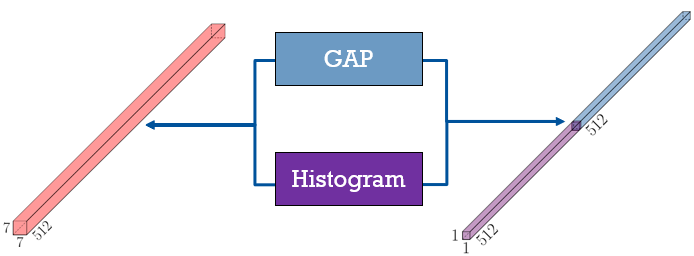}\label{fig:parallel}}
			\subcaptionbox{Series}{\includegraphics[width=\hsize]{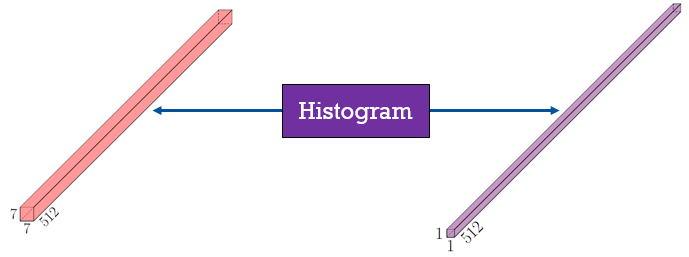}\label{fig:series}}  
		}
	}
    	\caption{Convolutional backbone and histogram model configurations used for multi-site dataset experiments. Resnet18 was used as the backbone architecture to extract convolutional features. In Figure \ref{fig:model_config}a, the convolutional, ReLU, and max pooling layers are represented with the light yellow, orange, and purple colors respectively. The local features maps produced by the histogram layer are $2 \times 2 \times 128$. The histogram features are reshaped to be the same size (\textit{i.e.}, $1 \times 1 \times 512$) as the feature vector produced by the global average pooling layer (GAP).}
    	\label{fig:model_config}
\end{figure*}
We implemented a) shallow and b) deep networks to evaluate the statistical and structural features of the SAS imagery. Following \cite{peeples2021histogram}, the shallow models were composed of a feature extractor (histogram or convolutional layer), global average pooling layer (GAP), and output classification layer. The backbone architecture for the deep networks was Resnet18 (Figure \ref{fig:model_config}a). In this work, we investigated series and parallel configuration (previously used in \cite{peeples2021histogram,wang2016learnable}) as shown in Figures \ref{fig:model_config}b and \ref{fig:model_config}c respectively. The series configuration would encode only statistical texture information before the output classification layer. The parallel configuration would consist of both statistical and structural texture features. As a result, we are able to observe the impact of structural, statistical, and combination of these texture features through the performance of the baseline (\textit{i.e.}, ResNet18), series histogram, and parallel histogram models respectively for SAS data. 

\section{Experimental Procedure}

\subsection{Dataset Description}

\paragraph{Statistical PISAS Dataset}
	\begin{figure}[htb]
	\centering 
	\begin{subfigure}{.15\textwidth}{
			\includegraphics[draft=false,width=\textwidth]{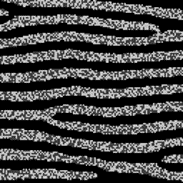}
			\caption{T1\_S1}
			\label{fig:binomialSR}
		}
	\end{subfigure}
	\begin{subfigure}{.15\textwidth}{
			\includegraphics[draft=false,width=\textwidth]{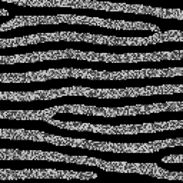}
			\caption{T1\_S2}
			\label{fig:multinomialSR}
		}
	\end{subfigure}
	\begin{subfigure}{.15\textwidth}{
			\includegraphics[draft=false,width=\textwidth]{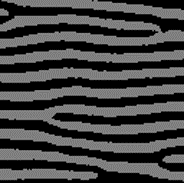}
			\caption{T1\_S3}
			\label{fig:constantSR}
		}
	\end{subfigure}

	\begin{subfigure}{.15\textwidth}{
		\includegraphics[draft=false,width=\textwidth]{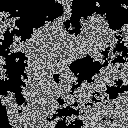}
		\caption{T10\_S1}
		\label{fig:binomialRocky}
	}
	\end{subfigure}
	\begin{subfigure}{.15\textwidth}{
			\includegraphics[draft=false,width=\textwidth]{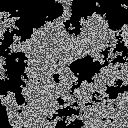}
			\caption{T10\_S2}
			\label{fig:multinomialRocky}
		}
	\end{subfigure}
	\begin{subfigure}{.15\textwidth}{
			\includegraphics[draft=false,width=\textwidth]{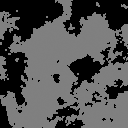}
			\caption{T10\_S3}
			\label{fig:constantRocky}
		}
	\end{subfigure}
	\caption[PISAS Sand Ripple Images]{Example of sand ripple (``T1") and rocky (``T10") images from the PISAS dataset with different statistical distributions. The same statistical classes (multinomial, binomial, and constant) and two SAS-related structures (\textit{i.e.}, sand ripple and rocky) were used to generate a total of six distinct classes. The binomial, multinomial, and constant textures are referenced as ``S1", ``S2", and ``S3" respectively. The subset of PISAS consisted of a total of $14,208$ images.}
	\centering \label{fig:PISAS_SR}
\end{figure}

The Pseudo Image SAS (PISAS) dataset \cite{stewart2021image} contains ten classes of sand ripple and rocky textures. Instead of the simple structures of cross, checkerboard, and stripe as shown in Figure \ref{fig:textures}, one class of sand ripple and rocky textures were selected to represent more SAS-related structures. Examples of sand ripple and rocky textures in the three different statistical classes are shown in Figure \ref{fig:PISAS_SR}. The statistical PISAS dataset consisted of 14,208 images.

\paragraph{Multi-site Dataset}
	\begin{figure}[htb]
	\centering 
	\begin{subfigure}{.117\textwidth}{
			\includegraphics[draft=false,width=\textwidth]{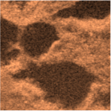}
			\caption{Craters}
			\label{fig:craters}
		}
	\end{subfigure}
	\begin{subfigure}{.117\textwidth}{
			\includegraphics[draft=false,width=\textwidth]{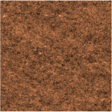}
			\caption{Flat}
			\label{fig:flat}
		}
	\end{subfigure}
	\begin{subfigure}{.117\textwidth}{
			\includegraphics[draft=false,width=\textwidth]{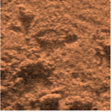}
			\caption{Rocky}
			\label{fig:rocky}
		}
	\end{subfigure}
	\begin{subfigure}{.117\textwidth}{
			\includegraphics[draft=false,width=\textwidth]{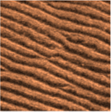}
			\caption{Sandripple}
			\label{fig:sandripple}
		}
	\end{subfigure}
	\caption{Example of images from multi-site dataset \cite{walker2021explainable}. The dataset consisted of real-world SAS imagery pure texture regions from multiple locations on the seafloor. Four seafloor textures types were identified to train, validate and test the model: craters, flat, rocky, and sand ripple. The multi-site dataset contained 118 total images.}
	\centering \label{fig:multi-site}
\end{figure}

Example images for the multi-site dataset \cite{walker2021explainable} are shown in Figure \ref{fig:multi-site}. The dataset consisted of 94 training, 12 validation, and 12 test images (118 images total) of pure textures from multiple locations along the seafloor. Random patches were sampled from the validation and test splits five times to generate 60 images for each set. A total of four classes comprised the dataset: flat, rocky, sand ripple, and craters. 

\subsection{Experimental Design}

\paragraph{Shallow Networks}
 We followed a similar training procedure as in \cite{peeples2021histogram} for the synthetic dataset experiments where each model was evaluated as a \textit{local} feature extractor. The network architecture consisted of a feature extractor (convolutional or histogram layer), global average pooling layer, and fully connected output layer. The ``CNN" model has a convolutional layer with ReLU as the feature extractor while the ``Hist" model used the histogram layer as the feature extractor. The experimental settings for each model was the following: 70/10/20 training, validation, and testing splits, 300 epochs, and early stopping by checking if the validation loss did not decrease after 10 consecutive epochs. The input image
 
   \begin{figure*}[htb]
	\centering 
	\begin{subfigure}{.495\textwidth}{
			\includegraphics[draft=false,width=\textwidth]{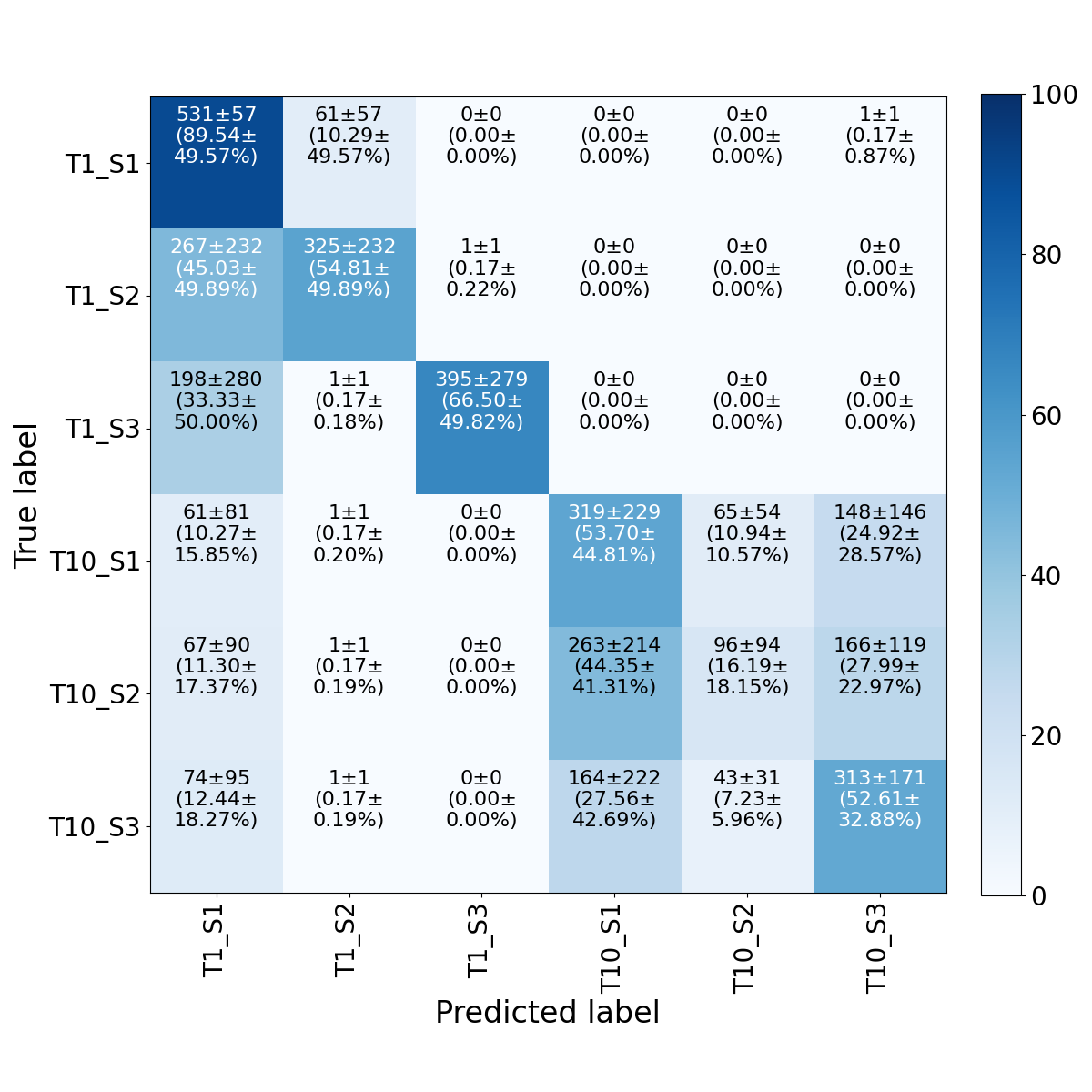}
			\caption{CNN ($55.65 \pm 21.0$)}
			\label{fig:PISAS_CNN_CM}
		}
	\end{subfigure}
	\begin{subfigure}{.495\textwidth}{
			\includegraphics[draft=false,width=\textwidth]{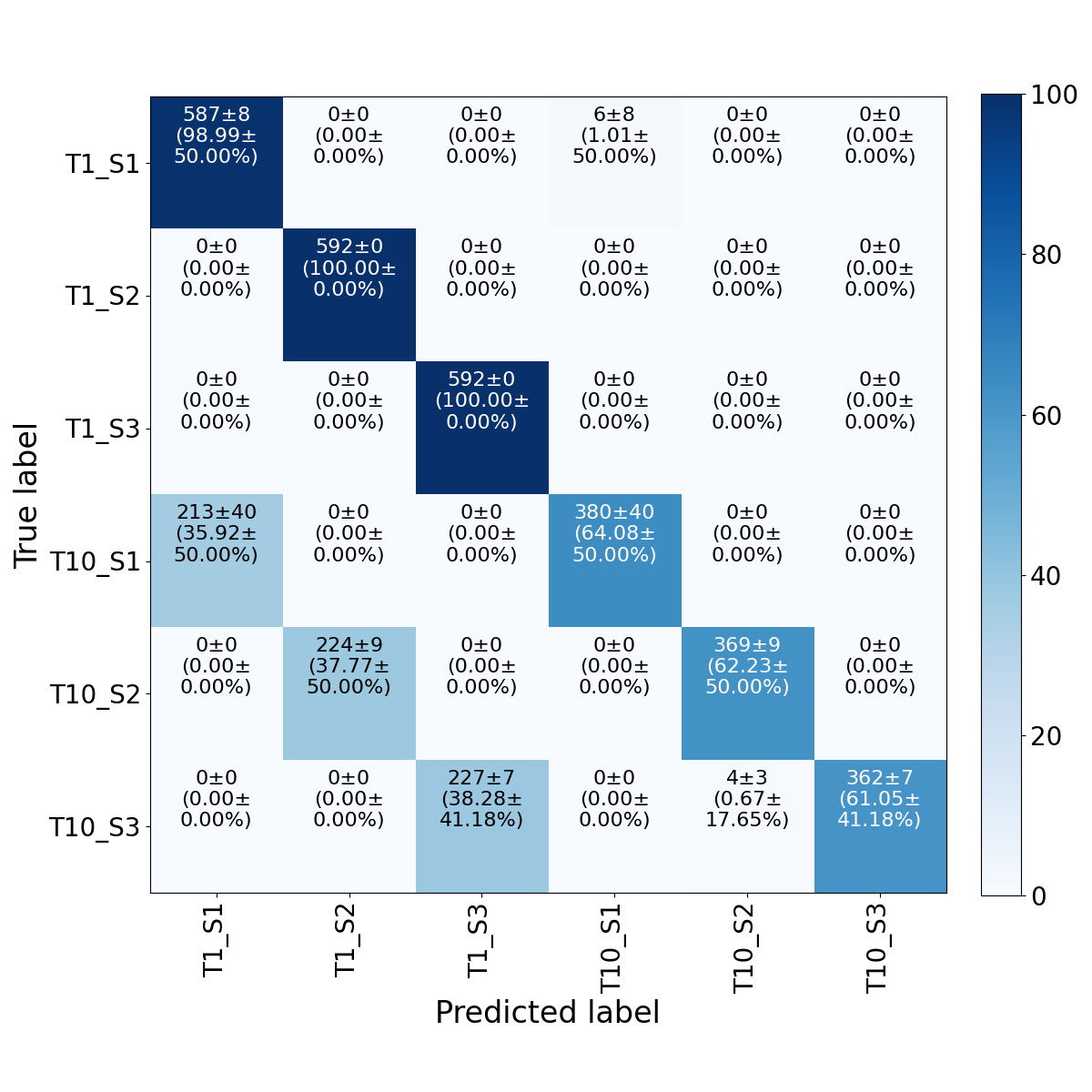}
			\caption{Hist ($81.11 \pm 0.76$)}
			\label{fig:PISAS_RBF_CM}
		}
	\end{subfigure}
	\caption{Average confusion matrices for the convolutional and histogram layer models on the PISAS dataset. ``T1" and ``T10" are sand ripple and rocky textures respectively. ``S1", ``S2", and ``S3" are the binomial, multinomial, and constant distributions respectively.}
	\centering \label{fig:PISAS_CM}
    \end{figure*}
 
 \noindent patches were $128 \times 128$. The model parameters were adjusted so that the kernel size was $7 \times 7$ with a $3 \times 3$ stride, and a batch size of $128$ was used to train each model. The number of bins for the RBF histogram layer model (\textit{i.e.}, Hist) and convolutional kernels for the CNN model were set to three. The histogram layer bins were initialized to be equally spaced between the range of the input data (\textit{i.e.}, $0$ and $1$). A total of three experimental runs were performed.

\paragraph{Deep Networks}
The HistRes\_$B$ model \cite{peeples2021histogram} was applied to the multi-site dataset, and two configurations were investigated: parallel and series. The training details were the following: 100 epochs, learning rate of $0.001$, Adam optimization \cite{kingma2015adam}, mini batch size of 16, $224 \times 224$ input image patches, and random crop data augmentation. A total of five experimental runs of random initialization were completed to evaluate the stability of each method. The number of bins was varied from $4$, $8$, and $16$.

\subsection{Results and Discussion}
\label{sec:parallelvseries}

\begin{figure*}[htb]
	\centering 
	\begin{subfigure}{.495\textwidth}{
			\includegraphics[draft=false,width=\textwidth]{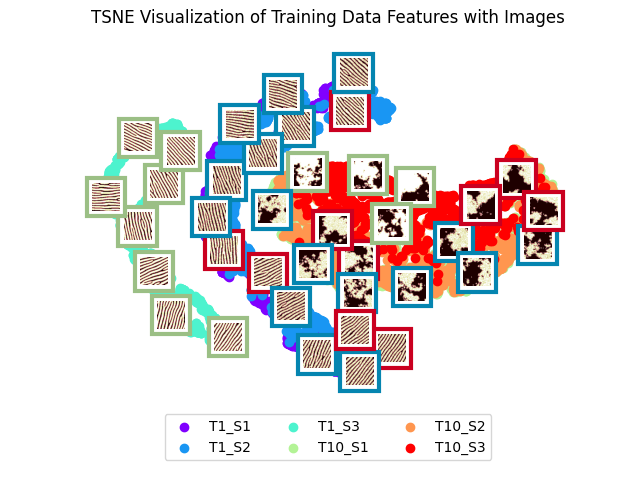}
			\caption{CNN ($77.70$)}
			\label{fig:PISAS_CNN_TSNE}
		}
	\end{subfigure} 
	\begin{subfigure}{.495\textwidth}{
			\includegraphics[draft=false,width=\textwidth]{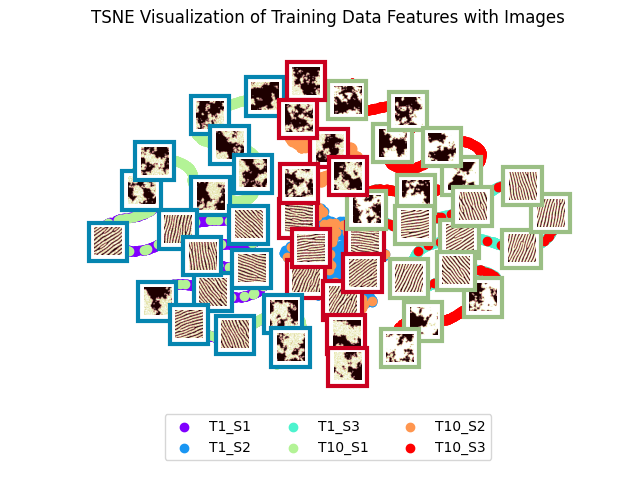}
			\caption{Hist ($82.18$)}
			\label{fig:PISAS_RBF_TSNE}
		}
	\end{subfigure}
	\caption[PISAS t-SNE Projections]{t-SNE projections for the convolutional and histogram layer models on the PISAS dataset. The test classification accuracy is shown in parenthesis. Example images from each class are shown. The frames around each image represent the statistical class of the PISAS dataset. Red, green, and blue frames represent the multinomial, constant, and binomial distributions respectively. The CNN model appears to map structural textures near one another in the projected space (\textit{i.e.}, sand ripple images are grouped to the left region of the projection and the rocky textures are grouped in the right of the projection). On the other hand, the histogram layer model maps statistical textures close to one another. The binomial statistical textures (blue frames) are grouped in the left of the projection. The multinomial (red frames) and constant (green frames) statistical textures are clustered in the center and right of the projection respectively.}
	\centering \label{fig:PISAS_TSNE}
\end{figure*}

\begin{figure*}[htb!]
	\centering 
	\begin{subfigure}{.32\textwidth}{
			\includegraphics[draft=false,width=\textwidth]{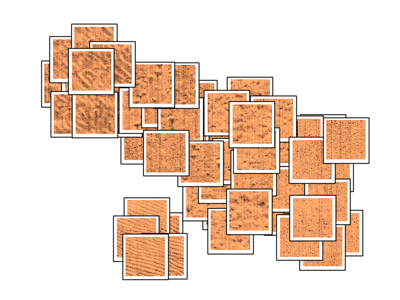}
			\caption{ResNet18 ($95.00$)}
			\label{fig:SAS_ResNet18}
		}
	\end{subfigure}
	\begin{subfigure}{.32\textwidth}{
			\includegraphics[draft=false,width=\textwidth]{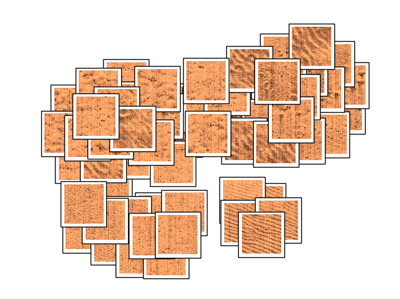}
			\caption{Series ($86.67$)}
			\label{fig:SAS_HistRes16Series}
		}
	\end{subfigure}
	\begin{subfigure}{.28\textwidth}{
			\includegraphics[draft=false,width=\textwidth]{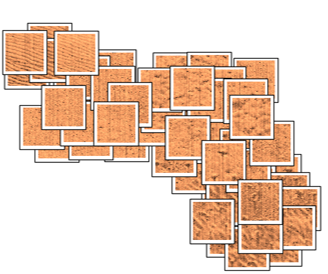}
			\caption{Parallel ($100.00$)}
			\label{fig:SAS_HistRes16Parallel}
		}
	\end{subfigure}
	\caption[SAS t-SNE Projections]{t-SNE projections for the best baseline ResNet18, HistRes\_16 Series, and HistRes\_16 Parallel. The test classification accuracy for the model is shown in parenthesis. The baseline ResNet18 did not have the histogram layer; as a result, the model will focus on structural textures. The projection of the ResNet18 features on the test data highlight that images with similar structures are in the mapped near one another in the lower dimensional space. Sand ripple images are to the left of the projection while there is a transition to rocky and crater textures in the middle with flat textures projected to the far right. The contrary is shown for the Series histogram model. Textures with similar statistics (\textit{i.e.}, pixel intensities) are projected near one another. The parallel model consist of both statistical and structural textures information; therefore, we observe a mix of the different textures in the projected space.}
	\centering \label{fig:SAS_TSNE}
\end{figure*}

\paragraph{CNN vs. Histogram Layer Models}
\begin{table}[htb!]
	\centering
	\caption[CNN and RBF on PISAS]{The average test accuracy for each model per class type is shown. The best average performance is bolded. The statistical labels relay the classification performance for predicting the distribution of the foreground pixels, while the structural labels show each model's prediction for sand ripple or rocky textures. Three experimental runs were conducted, and error values are reported with $\pm1$ standard deviation.}
	\begin{tabular}{|c|c|c|}
		\hline
		Label & CNN & Hist \\ \hline
		Both & 55.65$\pm$21.0 & \textbf{81.11$\pm$0.76}\\ \hline
		Statistical & 57.37$\pm$18.8 & \textbf{99.90$\pm$0.07}\\ \hline
		Structural  & \textbf{94.28$\pm$7.51} & 81.21$\pm$0.73\\ \hline
	\end{tabular}
	\label{tab:PISAS}
\end{table}

\begin{table}[htb]
	\centering
	\caption{The average Calinski-Harabasz index \cite{calinski1974dendrite} for each model per class type is shown. The best average performance is bolded (larger values indicate better separability and compactness). The statistical labels demonstrate the feature quality for predicting the distribution of the foreground pixels, while the structural labels show the feature quality for distinguishing between the sand ripple or rocky textures. Three experimental runs were conducted, and error values are reported with $\pm1$ standard deviation.}
	\begin{tabular}{|c|c|c|}
		\hline
		Label & CNN & Hist \\ \hline
		Both & 3,333.88$\pm$2,163.06 & \textbf{26,129.02$\pm$236.77}\\ \hline
		Statistical & 40.55$\pm$32.50 & \textbf{50,516.00$\pm$460.19}\\ \hline
		Structural  & \textbf{15,399.12$\pm$9951.00} & 134.28$\pm$3.92\\ \hline
	\end{tabular}
	\label{tab:Cluster}
\end{table}	 

The results of each shallow network on the PISAS test data are shown in Table \ref{tab:PISAS}. Similar trends in the results follow for the SAS structures as observed by the simple structures in \cite{peeples2021histogram}. The convolutional model can easily distinguish the sand ripple and rocky textures, but struggles with the statistical differences in comparison to the histogram layer. The histogram layer can be important in SAS application when the statistics of the data may change due to environmental conditions such as speckle noise \cite{abu2018robust}. The confusion matrices also highlight the differences between the two models, as shown in Figure \ref{fig:PISAS_CM}. The histogram layer model has the most difficulty with the structural changes between the sand ripple and rocky images, since there is equal weighting on spatial contribution of neighbors due to the average pooling operation. However, the CNN has more degradation in performance when observing statistical changes, particularly for the rocky texture (``T10").

Interesting observations can also be seen qualitatively in the t-SNE projection of the training data, shown in Figure \ref{fig:PISAS_TSNE}. The t-SNE projections used the same initialization and random seed. The only difference is the features that were projected into the lower dimensional space (convolutional or histogram features). For the CNN, the sand ripple and rocky images are projected near each other. However, the statistical textures are mapped to similar regions of the feature space after the t-SNE projection. The clusters are also appeared tighter for the histogram layer model, indicating that the histogram layer may be able to handle small intra-class variations in the data through the soft binning operations to characterize images in the same class.

Additional analysis was needed to verify the plausible conclusions discussed above to ensure the difference is a result of the higher dimensional structure and not an artifact of the 

\begin{figure*}[htb!]
	\centering
	\includegraphics[width=.92\linewidth]{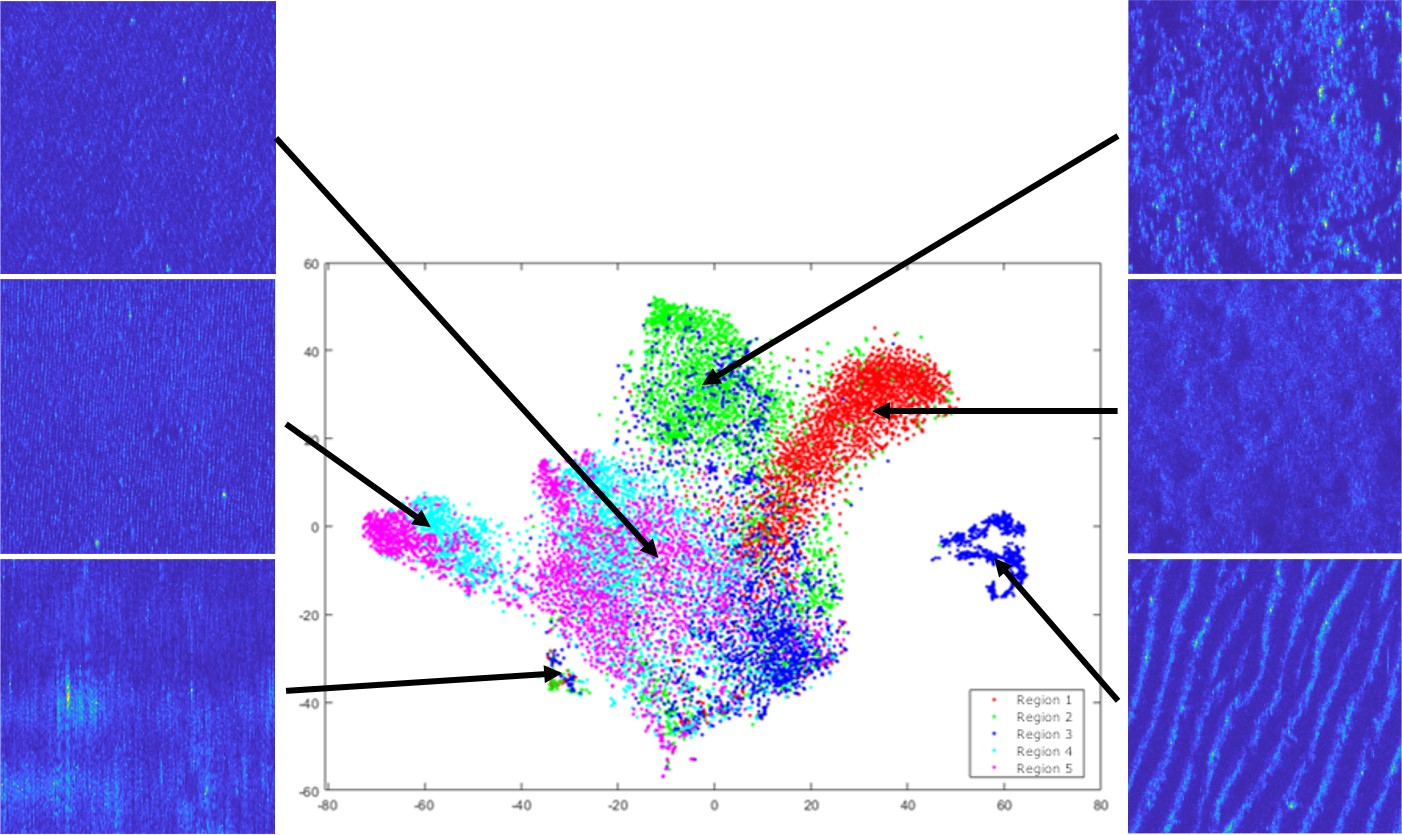}
	\caption[Histogram t-SNE Embedding]{t-SNE projections of SAS images from various locations of the seafloor passed through the best parallel HistRes\_16 model. The colors indicate images collected from different regions. The best parallel model was trained on a small subset of labelled SAS imagery (\textit{i.e.}, multi-site dataset). The model was able to learn features that generalized to new images from different regions. In the projected space, we see meaningful groupings such as sand ripple textures in the far right of the projection. The model was also able to extract features that grouped images containing cross talk, as shown in the bottom left-hand corner of the projection.}
	\label{fig:Regions_TSNE}
\end{figure*}

\noindent t-SNE mapping. Therefore, we used a cluster validity metric to assess the quality of the features learned by each method with regard to the labels. The Calinski-Harabsz index \cite{calinski1974dendrite} considers both intra- and inter-cluster distances. Larger values indicate more compact and separated classes in the feature space. We observed that the metrics matched our analysis of the lower dimensional space, as shown in Table \ref{tab:Cluster}. The histogram layer model outperformed the CNN for statistical textures, while the CNN was superior for structural textures. Overall, we noted that the histogram features had the best features across all classes. Experiments on additional SAS images in the \textit{Parallel vs. Series Configuration} section further demonstrated the power of statistical features for deep learning models.

\paragraph{Parallel vs. Series Configuration}

\begin{table}[t]
	\centering
	\caption{The average test accuracy for each model on the SAS dataset is shown. The best average performance is bolded. The baseline ResNet18 model achieved a test classification accuracy of 84.66$\pm$11.42. Five experimental runs were conducted, and error values are reported with $\pm1$ standard deviation.}
	\begin{tabular}{|c|c|c|c|}
		\hline
		Configuration & HistRes\_4 & HistRes\_8 & HistRes\_16 \\ \hline
		Parallel      & \textbf{90.33$\pm$4.00} & \textbf{82.67$\pm$8.27}                      & \textbf{88.33$\pm$9.60}  \\ \hline
		Series        & 
		77.33$\pm$5.01 & 
		73.33$\pm$5.06 &
		81.00$\pm$4.03        \\ \hline
	\end{tabular}
	\label{tab:SAS}
\end{table}

The test performance on the multi-site dataset for the model configuration experiments is shown in Table \ref{tab:SAS}. The parallel and series models were robust to the number of bins, as there was not a significant difference in performance (\textit{i.e.}, overlapping error bars). The parallel model outperformed both the baseline and series models. This further validated the utility of using both statistical and structural information to maximize texture classification. In addition to the overall test performance, a qualitative analysis was performed using t-SNE (Figure \ref{fig:SAS_TSNE}). The convolutional model mapped textures with similar structures near each other, while the histogram models mapped points with similar statistics near one another. For the parallel configuration, there was a mix of similar statistical and structural SAS images near one another in the projected space.

Once training and evaluation was completed on the subset of labeled SAS images, the model was applied to hold-out SAS images from multiple locations on the seafloor. The t-SNE projection is shown in Figure \ref{fig:Regions_TSNE}. The model successfully extracted features from the hold-out images that showed meaningful groupings in the lower dimensional projection space. The parallel HistRes\_16 model can also potentially be used to identify images with cross-talk (bottom left-hand corner of Figure \ref{fig:Regions_TSNE}). The results in this section demonstrated the use of the histogram layer for various real-world applications for SAS texture classification.

\section{Conclusion}
We presented a novel application of histogram layer(s) for SAS image classification. The results presented in this work show the impact of statistical texture features within deep learning frameworks. For SAS imagery, histogram layer(s) provide a powerful feature representation that can jointly be tuned with CNNs to optimize performance. Future work includes exploring other convolutional backbones, investigating other binning functions, performing semantic segmentation, and extensive comparisons with other SAS image analysis approaches.

\balance
\bibliographystyle{IEEEtran}
\bibliography{refs.bib}

\begin{thebibliography}{10}
\providecommand{\url}[1]{#1}
\csname url@samestyle\endcsname
\providecommand{\newblock}{\relax}
\providecommand{\bibinfo}[2]{#2}
\providecommand{\BIBentrySTDinterwordspacing}{\spaceskip=0pt\relax}
\providecommand{\BIBentryALTinterwordstretchfactor}{4}
\providecommand{\BIBentryALTinterwordspacing}{\spaceskip=\fontdimen2\font plus
\BIBentryALTinterwordstretchfactor\fontdimen3\font minus
  \fontdimen4\font\relax}
\providecommand{\BIBforeignlanguage}[2]{{%
\expandafter\ifx\csname l@#1\endcsname\relax
\typeout{** WARNING: IEEEtran.bst: No hyphenation pattern has been}%
\typeout{** loaded for the language `#1'. Using the pattern for}%
\typeout{** the default language instead.}%
\else
\language=\csname l@#1\endcsname
\fi
#2}}
\providecommand{\BIBdecl}{\relax}
\BIBdecl

\bibitem{hayes2009synthetic}
M.~P. Hayes and P.~T. Gough, ``Synthetic aperture sonar: a review of current
  status,'' \emph{IEEE journal of oceanic engineering}, vol.~34, no.~3, pp.
  207--224, 2009.

\bibitem{richard2021deep}
G.~Richard, J.-M. Le~Caillec, J.~Habonneau, and D.~Gueriot, ``A deep sas atr
  explainability framework assessment,'' in \emph{OCEANS 2021: San
  Diego--Porto}.\hskip 1em plus 0.5em minus 0.4em\relax IEEE, 2021, pp. 1--5.

\bibitem{sun2022iterative}
Y.-C. Sun, I.~D. Gerg, and V.~Monga, ``Iterative, deep synthetic aperture sonar
  image segmentation,'' \emph{IEEE Transactions on Geoscience and Remote
  Sensing}, vol.~60, pp. 1--15, 2022.

\bibitem{stewart2021image}
D.~Stewart, S.~Johnson, and A.~Zare, ``Image-to-height domain translation for
  synthetic aperture sonar,'' \emph{arXiv preprint arXiv:2112.06307}, 2021.

\bibitem{williams2009unsupervised}
D.~P. Williams, ``Unsupervised seabed segmentation of synthetic aperture sonar
  imagery via wavelet features and spectral clustering,'' in \emph{2009 16th
  IEEE International Conference on Image Processing (ICIP)}.\hskip 1em plus
  0.5em minus 0.4em\relax IEEE, 2009, pp. 557--560.

\bibitem{kohntopp2017seafloor}
D.~Kohntopp, B.~Lehmann, D.~Kraus, and A.~Birk, ``Seafloor classification for
  mine countermeasures operations using synthetic aperture sonar images,'' in
  \emph{OCEANS 2017-Aberdeen}.\hskip 1em plus 0.5em minus 0.4em\relax IEEE,
  2017, pp. 1--5.

\bibitem{geirhos2018imagenet}
R.~Geirhos, P.~Rubisch, C.~Michaelis, M.~Bethge, F.~A. Wichmann, and
  W.~Brendel, ``Imagenet-trained cnns are biased towards texture; increasing
  shape bias improves accuracy and robustness,'' in \emph{International
  Conference on Learning Representations}, 2018.

\bibitem{hermann2020origins}
K.~Hermann, T.~Chen, and S.~Kornblith, ``The origins and prevalence of texture
  bias in convolutional neural networks,'' \emph{Advances in Neural Information
  Processing Systems}, vol.~33, pp. 19\,000--19\,015, 2020.

\bibitem{peeples2021histogram}
J.~Peeples, W.~Xu, and A.~Zare, ``Histogram layers for texture analysis,''
  \emph{IEEE Transactions on Artificial Intelligence}, vol.~3, no.~4, pp.
  541--552, 2022.

\bibitem{zhu2021learning}
L.~Zhu, D.~Ji, S.~Zhu, W.~Gan, W.~Wu, and J.~Yan, ``Learning statistical
  texture for semantic segmentation,'' in \emph{Proceedings of the IEEE/CVF
  Conference on Computer Vision and Pattern Recognition}, 2021, pp.
  12\,537--12\,546.

\bibitem{liu2019bow}
L.~Liu, J.~Chen, P.~Fieguth, G.~Zhao, R.~Chellappa, and M.~Pietik{\"a}inen,
  ``From bow to cnn: Two decades of texture representation for texture
  classification,'' \emph{International Journal of Computer Vision}, vol. 127,
  no.~1, pp. 74--109, 2019.

\bibitem{peeples2019comparison}
J.~Peeples, M.~Cook, D.~Suen, A.~Zare, and J.~Keller, ``Comparison of
  possibilistic fuzzy local information c-means and possibilistic k-nearest
  neighbors for synthetic aperture sonar image segmentation,'' in
  \emph{Detection and Sensing of Mines, Explosive Objects, and Obscured Targets
  XXIV}, vol. 11012.\hskip 1em plus 0.5em minus 0.4em\relax International
  Society for Optics and Photonics, 2019, p. 110120T.

\bibitem{zare2017possibilistic}
A.~Zare, N.~Young, D.~Suen, T.~Nabelek, A.~Galusha, and J.~Keller,
  ``Possibilistic fuzzy local information c-means for sonar image
  segmentation,'' in \emph{2017 IEEE Symposium Series on Computational
  Intelligence (SSCI)}.\hskip 1em plus 0.5em minus 0.4em\relax IEEE, 2017, pp.
  1--8.

\bibitem{williams2015fast}
D.~P. Williams, ``Fast unsupervised seafloor characterization in sonar imagery
  using lacunarity,'' \emph{IEEE transactions on Geoscience and Remote
  Sensing}, vol.~53, no.~11, pp. 6022--6034, 2015.

\bibitem{haralick1973textural}
R.~M. Haralick, K.~Shanmugam \emph{et~al.}, ``Textural features for image
  classification,'' \emph{IEEE Transactions on systems, man, and cybernetics},
  no.~6, pp. 610--621, 1973.

\bibitem{zhu2014model}
Z.~Zhu, X.~Xu, L.~Yang, H.~Yan, S.~Peng, and J.~Xu, ``A model-based sonar image
  atr method based on sift features,'' in \emph{OCEANS 2014-TAIPEI}.\hskip 1em
  plus 0.5em minus 0.4em\relax IEEE, 2014, pp. 1--4.

\bibitem{cobb2010parametric}
J.~T. Cobb, K.~C. Slatton, and G.~J. Dobeck, ``A parametric model for
  characterizing seabed textures in synthetic aperture sonar images,''
  \emph{IEEE Journal of Oceanic Engineering}, vol.~35, no.~2, pp. 250--266,
  2010.

\bibitem{sedighi2017histogram}
V.~Sedighi and J.~Fridrich, ``Histogram layer, moving convolutional neural
  networks towards feature-based steganalysis,'' \emph{Electronic Imaging},
  vol. 2017, no.~7, pp. 50--55, 2017.

\bibitem{wang2016learnable}
Z.~Wang, H.~Li, W.~Ouyang, and X.~Wang, ``Learnable histogram: Statistical
  context features for deep neural networks,'' in \emph{European Conference on
  Computer Vision}.\hskip 1em plus 0.5em minus 0.4em\relax Springer, 2016, pp.
  246--262.

\bibitem{yusuf2020differentiable}
I.~Yusuf, G.~Igwegbe, and O.~Azeez, ``Differentiable histogram with
  hard-binning,'' \emph{arXiv preprint arXiv:2012.06311}, 2020.

\bibitem{sadeghi2022histnet}
H.~Sadeghi and A.-A. Raie, ``Histnet: Histogram-based convolutional neural
  network with chi-squared deep metric learning for facial expression
  recognition,'' \emph{Information Sciences}, vol. 608, pp. 472--488, 2022.

\bibitem{materka1998texture}
A.~Materka, M.~Strzelecki \emph{et~al.}, ``Texture analysis methods--a
  review,'' \emph{Technical university of lodz, institute of electronics, COST
  B11 report, Brussels}, pp. 9--11, 1998.

\bibitem{humeau2019texture}
A.~Humeau-Heurtier, ``Texture feature extraction methods: A survey,''
  \emph{IEEE Access}, vol.~7, pp. 8975--9000, 2019.

\bibitem{weszka1976comparative}
J.~S. Weszka, C.~R. Dyer, and A.~Rosenfeld, ``A comparative study of texture
  measures for terrain classification,'' \emph{IEEE transactions on Systems,
  Man, and Cybernetics}, no.~4, pp. 269--285, 1976.

\bibitem{haralick1979statistical}
R.~M. Haralick \emph{et~al.}, ``Statistical and structural approaches to
  texture,'' \emph{Proceedings of the IEEE}, vol.~67, no.~5, pp. 786--804,
  1979.

\bibitem{ojala1996comparative}
T.~Ojala, M.~Pietik{\"a}inen, and D.~Harwood, ``A comparative study of texture
  measures with classification based on featured distributions,'' \emph{Pattern
  recognition}, vol.~29, no.~1, pp. 51--59, 1996.

\bibitem{walker2021explainable}
S.~Walker, J.~Peeples, J.~Dale, J.~Keller, and A.~Zare, ``Explainable
  systematic analysis for synthetic aperture sonar imagery,'' in \emph{2021
  IEEE International Geoscience and Remote Sensing Symposium IGARSS}, 2021, pp.
  2835--2838.

\bibitem{kingma2015adam}
D.~P. Kingma and J.~Ba, ``Adam: A method for stochastic optimization,'' in
  \emph{International Conference on Learning Representations (ICLR)}, 2015.

\bibitem{calinski1974dendrite}
T.~Cali{\'n}ski and J.~Harabasz, ``A dendrite method for cluster analysis,''
  \emph{Communications in Statistics-theory and Methods}, vol.~3, no.~1, pp.
  1--27, 1974.

\bibitem{abu2018robust}
A.~Abu and R.~Diamant, ``Robust image denoising for sonar imagery,'' in
  \emph{2018 OCEANS-MTS/IEEE Kobe Techno-Oceans (OTO)}.\hskip 1em plus 0.5em
  minus 0.4em\relax IEEE, 2018, pp. 1--5.

\end{thebibliography}

\end{document}